\documentclass[sn-mathphys-num]{sn-jnl}

\usepackage{graphicx}%
\usepackage{multirow}%
\usepackage{amsmath,amssymb,amsfonts}%
\usepackage{amsthm}%
\usepackage{mathrsfs}%
\usepackage[title]{appendix}%
\usepackage{xcolor}%
\usepackage{textcomp}%
\usepackage{manyfoot}%
\usepackage{booktabs}%
\usepackage{algorithm}%
\usepackage{algorithmicx}%
\usepackage{algpseudocode}%
\usepackage{listings}%
\usepackage{cleveref}

\usepackage[utf8]{inputenc} 
\usepackage[T1]{fontenc}    
\usepackage{hyperref}       
\usepackage{url}            
\usepackage{booktabs}       
\usepackage{amsfonts}       
\usepackage{nicefrac}       
\usepackage{microtype}      
\usepackage{xcolor}         

\theoremstyle{thmstyleone}%
\theoremstyle{thmstyletwo}%

\theoremstyle{thmstylethree}%

\begin{document}

\title[Article Title]{Anonymization Prompt Learning for Facial Privacy-Preserving Text-to-Image Generation}


\author[1,2]{\fnm{Liang} \sur{Shi}}\email{liang.shi@vipl.ict.ac.cn}

\author*[1,2]{\fnm{Jie} \sur{Zhang}}\email{zhangjie@ict.ac.cn}

\author[1,2]{\fnm{Shiguang} \sur{Shan}}\email{sgshan@ict.ac.cn}

\affil[1]{\orgdiv{Institute of Computing Technology}, \orgname{Chinese Academy of Sciences}, \orgaddress{\city{Beijing}, \postcode{100190}, \country{China}}}

\affil[2]{\orgname{University of Chinese Academy of Sciences}, \orgaddress{\city{Beijing}, \postcode{100190}, \country{China}}}


\abstract{Text-to-image diffusion models, such as Stable Diffusion, generate highly realistic images from text descriptions. However, the generation of certain content at such high quality raises concerns. A prominent issue is the accurate depiction of identifiable facial images, which could lead to malicious deepfake generation and privacy violations. In this paper, we propose Anonymization Prompt Learning (APL) to address this problem. Specifically, we train a learnable prompt prefix for text-to-image diffusion models, which forces the model to generate anonymized facial identities, even when prompted to produce images of specific individuals. Extensive quantitative and qualitative experiments demonstrate the successful anonymization performance of APL, which anonymizes any specific individuals without compromising the quality of non-identity-specific image generation. Furthermore, we reveal the plug-and-play property of the learned prompt prefix, enabling its effective application across different pretrained text-to-image models for transferrable privacy and security protection against the risks of deepfakes.}




\maketitle

\section{Introduction}
\label{sec:intro}

The introduction of text-to-image diffusion models~\cite{ldm, dalle3, imagen} marks a significant milestone in the field of image generation. These models enable the creation of images known for their high resolution, realism, and close alignment with text descriptions. This progress sets new standards for artificial content, and has led to online text-to-image services~\cite{dalle3} providing free access to the general public.

However, the ability to forge realistic content with such high degrees of freedom has raised wide social concerns. Among the various problems discussed, the generation of identifiable facial images~\cite{antidreambooth, hyperdreambooth, imma} draws particular attention, as such realistic deepfake generation could lead to violation of privacy. Multiple incidents have occurred recently where text-to-image models were used to create fake images of public figures, which were often indistinguishable from real photos without careful scrutiny. It is anticipated that future iterations of text-to-image models will only intensify the issue, empowering users to realistically create any individual in any setting solely based on textual descriptions.

In wake of this problem, online service providers develop content moderation systems to avoid undesired face generation. For example, OpenAI's online image generation system DALLE-3~\cite{dalle3} employs a large language model to reject the generation of specific individuals. While effective, such measures introduce extra steps in processing, raising the cost of maintaining a service. To avoid such complicated moderation schemes, researchers also probe into the possibility of removing undesired content from generative models with concept editing~\cite{forgetmenot, erase, ablating}. These methods aim at forcing the model to forget sensitive concepts like personal identities~\cite{forgetmenot, erase}. However, most existing methods are constrained to removing only a handful of concepts without severely damaging generation quality. This inability to process more concepts is potentially due to their modification of all or a large subset of network parameters. Consequently, they are insufficient for defending against deepfake image generation, which involves an unlimited number of facial identities.

\begin{figure}[t]
  \centering
  \includegraphics[width=1\linewidth]{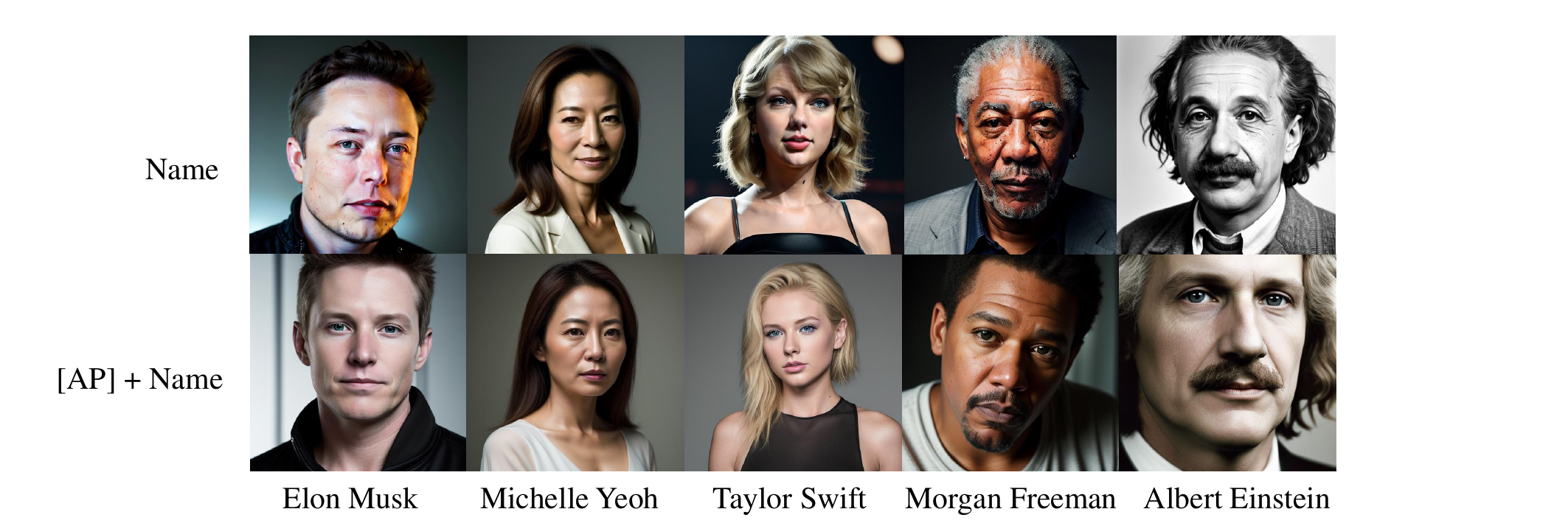}
  \caption{Images generated by Realistic Vision v5.1~\cite{realisticvision} when prompted with names of public figures. The base model generates highly convincing photos of given identities (top row), which may lead to malicious use. When inserting our Anonymization Prompt (AP) to the model (bottom row), the accuracies of generated identities are significantly reduced. }
  \label{fig1}
\end{figure}

In this paper, we propose Anonymization Prompt Learning (APL), a simple approach that addresses the deepfake generation issue. Unlike existing methods, APL achieves this without relying on additional content moderation modules or extensive editing of model parameters. To accomplish this, we introduce a learnable soft prompt prefix that is prepended to \textit{any} prompt input for text-to-image models. Given an identity-specific prompt, we train the prefix such that it forces the diffusion model into generating facial images with inaccurate identities, as demonstrated in~\cref{fig1}. This is achieved by denoising facial images so that the output aligns less with the requested identity, but closely with a prompt that describes the person's prominent facial attribute information. We further train the prefix to be inactive for prompts that do not specify certain identities by aligning the outputs with and without the prefix, thereby maintaining the model's generation quality. Collectively, a text-to-image diffusion model equipped with the Anonymization Prompt  functions almost identically to its original version, except for autonomously generating anonymized images when prompted to depict specific individuals. 

We quantitatively evaluate the anonymization performance with a face recognition model, which measures the distance between generated identities and the ground truths. Results show that our method facilitates the generation of identities significantly less similar to the ground truth identity. We establish that any identities can be anonymized, including those not seen during training. In addition, no significant degradation is observed on generation quality and text fidelity according to commonly-used metrics for evaluating image generation models. 

We further reveal several advantages of learning a prompt as opposed to training the entire model. Firstly, an Anonymization Prompt learned on one text-to-image diffusion model can be straightforwardly transferred to another model, and demonstrate competitive anonymization performance. This entails a plug-and-play property of our method, facilitating its applicability across a wide array of platforms and scenarios. Secondly, the prompt remains effective after model fine-tuning, as it can always be integrated to the model following the training process and continue to function. 

We summarize our contributions below:

(1) We propose Anonymization Prompt Learning (APL), a novel approach to mitigate the issue of deepfake generation by text-to-image models. This is achieved through a learnable soft prompt prefix that leads to the generation of anonymized images even for prompts specifying certain identities.

(2) We ensure the Anonymization Prompt does not affect the model's basic generative quality, so that online service providers can integrate the approach into their existing platforms without worrying about performance degradation. 

(3) Extensive experiments demonstrate the efficacy of the Anonymization Prompt in producing anonymized images when models are prompted for specific individuals, without degrading image quality or text fidelity for non-identity-specific requests. Furthermore, the Anonymization Prompt exhibits a plug-and-play property that allows it to be easily transferred and integrated into various text-to-image models. 

\section{Related Work}
\label{sec:related}

We present an overview of text-to-image diffusion models. The overview is followed by discussions of their potential misuse and implementations of content moderation practices. Finally, we briefly mention the field of face anonymization and contrast their task to ours in facial privacy protection. 

\subsection{Text-to-Image Diffusion Models}

The field of image generation has taken a leap forward thanks to the introduction of diffusion models~\cite{ddpm, ddim, adm}. Through iterative denoising steps, these models implement a mapping from random noise to the real image distribution, achieving superior performances in terms of image quality and diversity. More recently, Latent Diffusion Model (LDM) ~\cite{ldm} is proposed to operate the diffusion process in a compressed latent space. This practice reduces computational costs and enables the introduction of conditional inputs, such as text descriptions. Taking inspirations from LDM, current text-to-image diffusion models are celebrated for their ability to produce images of high resolution, realism, and close adherence to text descriptions. These advancements have resulted in huge successes like Stable Diffusion~\cite{ldm, sdxl}, Midjourney~\cite{midjourney}, and DALLE-3~\cite{dalle3}, which offer public access to either model parameters or through open API services. Such capabilities enable users to produce an extensive range of realistic creations tailored to their specific requests. 

While diffusion models enable numerous creative applications, their utilization raises wide societal concerns regarding potential malicious uses. Particularly, text-to-image models faithfully following textual inputs can be used for inappropriate generations. Among many concerns, one major issue is the generation of high-quality, identifiable facial images, which leads to concerning privacy issues. Several work~\cite{antidreambooth, hyperdreambooth, imma} confirm the capability of models to generate accurate identities when prompted with specific individuals. We target the face generation capability of these general-purposed text-to-image models, aiming to moderate them towards failure at identity-specific generation, but preserving abilities to generate high-quality non-identity-specific content. 

\subsection{Content Moderation for Text-to-Image Generation}

Online text-to-image service providers employ additional pre-processing or post-processing modules to reject generations deemed inappropriate, including images depicting specific individuals~\cite{dalle3}. Meanwhile, researchers of concept editing methods~\cite{erase, forgetmenot, ablating, uce} explore the modification of text-to-image models themselves to prevent the generation of designated concepts. ESD~\cite{erase} leverages a frozen teacher model to provide unconditional guidance for diffusion models towards forgetting of concepts. Concept Ablation~\cite{ablating} proposes model-based and noise-based ablation that both guide the predictions of diffusion models towards pre-defined anchor concepts. Forget-me-not~\cite{forgetmenot} modifies the cross-attention layers of diffusion models to minimize the attention scores between designated concepts and accurate generations. These methods primarily target a limited number of concepts to remove, and tend to suffer from catastrophic forgetting~\cite{uce} when the number of target concepts increases. This paper demonstrates that a learnable prompt prefix allows for the anonymization of an unlimited number of facial identities, without significantly impacting the quality of generations.

\subsection{Face Anonymization}
Face anonymization~\cite{blur, pixel, deepprivacy, ciagan, ldfa, attranon} strives to edit existing facial images to remove identity information, thereby protecting facial privacy. Earlier methods~\cite{blur, pixel} use blurring or pixelation to completely obscure faces. Recent approaches~\cite{deepprivacy, ciagan, ldfa, attranon} utilize generative models to replace existing faces with others that have different identities. Deepprivacy~\cite{deepprivacy} and CIAGAN~\cite{ciagan} use Generative Adversarial Networks~\cite{gan} conditioned on face landmarks to guide accurate anonymizations. LDFA~\cite{ldfa} leverages the powerful inpainting capability of diffusion models to enhance face anonymization for self-driving applications. A recent work~\cite{attranon} proposes to preserve facial attribute information in anonymization with the guidance of a face attribute detection model. 

In essence, existing methods focus on editing an existing image towards anonymity. However, when applied to generated images as a post-processing step, they naturally incur higher costs for service providers and hamper generation quality. On the contrary, our method ensures that generative models directly produce anonymized images without risking quality degradation or extra costs.


\section{Method}
\label{sec:method}

\subsection{Overview}

We propose Anonymization Prompt Learning (APL), a simple procedure that prevents the generation of identity-specific images in text-to-image diffusion models. We train an Anonymization Prompt that, by default, is attached to any prompt inputs to diffusion models. The Anonymization Prompt is inactivate for most prompt inputs, but forces the diffusion model to generate anonymized images when prepended to prompts specifying facial identities. 

\begin{figure}[t]
  \centering
  \includegraphics[width=1\linewidth]{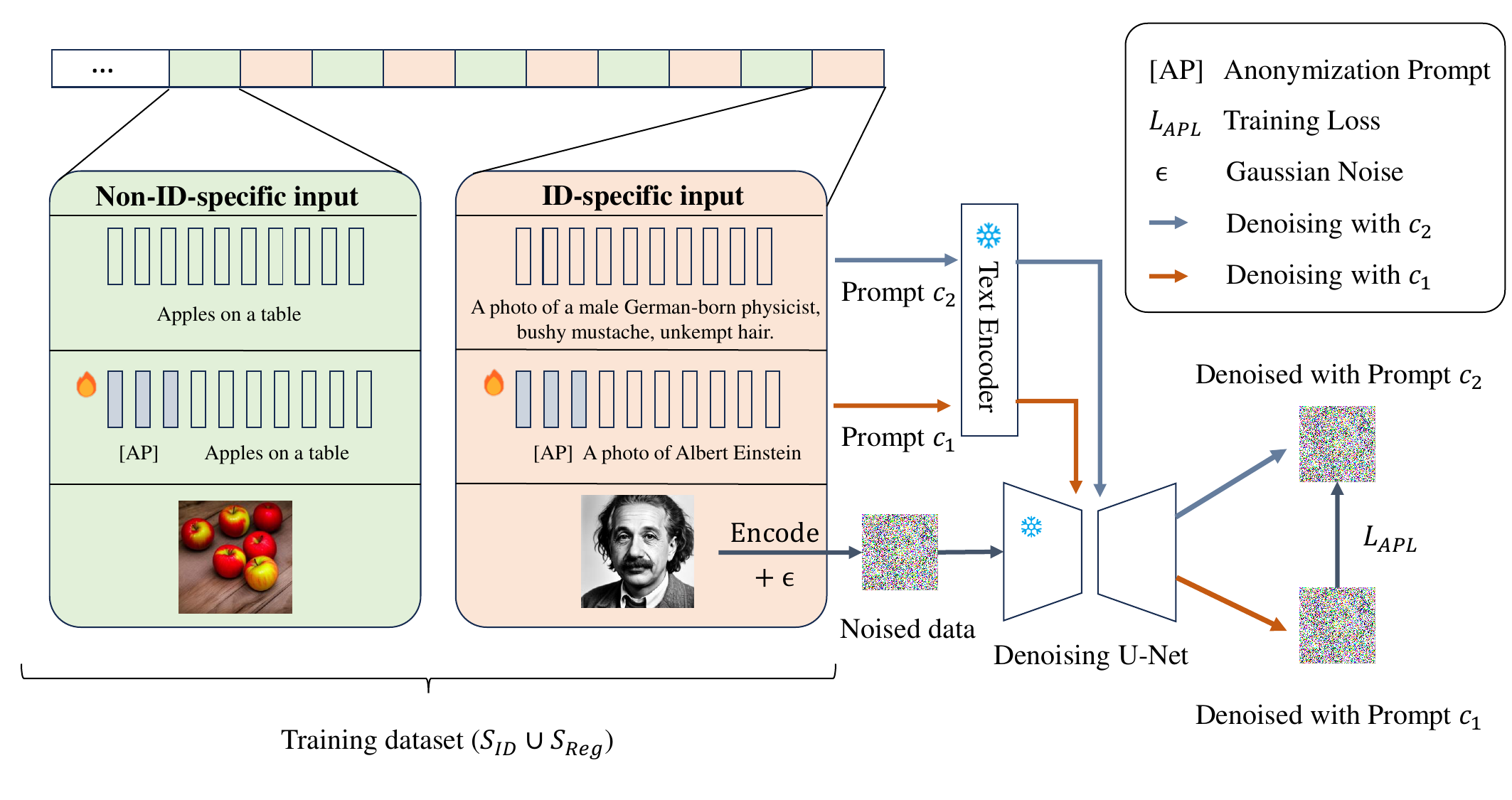}

  \caption{Pipeline of Anonymization Prompt Learning (APL). At any timestep, the diffusion model denoises images separately with two different prompts. For ID-specific inputs from $S_{ID}$, the Anonymization Prompt (AP) is prepended to prompts of names, and is trained to produce similar noise predictions to that of corresponding attribute descriptions. Similarly, For non-ID-specific inputs from $S_{Reg}$, the Anonymization Prompt is trained to be ineffective, avoiding generation quality degradation. Both the denoising U-Net and the text encoder are fixed during training. }
  \label{pipeline}

\end{figure}

In this section, we introduce the idea of learning to anonymize for diffusion models, and provide information on how to preserve prompt fidelity and generation quality of diffusion models with APL introduced. A detailed pipeline of APL is presented in~\cref{pipeline}. 

\vspace{-0.5em}
\subsection{Text-to-Image Diffusion Models}
\label{sec:t2i}

Diffusion models~\cite{ddpm} implement a mapping from random noise to the image distribution by performing a multi-step image denoising. This is achieved by training the model to perform image denoising at different time steps with different noise levels: 

\begin{equation}
L_{dm} = \mathbb{E}_{x_t, c, t \sim [1, T]} [||\epsilon_t - \epsilon_\theta (x_t, \tau(c), t) ||^2],
\end{equation}

where $\epsilon_t$ is a sampled Gaussian noise and $t$ is the denoising time step. In the context of Latent Diffusion Models~\cite{ldm}, $x_t$ is obtained by compressing real images into a latent space and subsequently adding noises. For text-to-image models, $c$ denotes the textual description guiding the image generation, and $\tau$ is implemented by a text encoder.

\subsection{Prompt Learning for Diffusion Models}
\label{sec:prompt}
In this paper, we achieve our target of anonymization through prompt learning. In specific, during the entire training process, the diffusion model $\epsilon_\theta$ as well as the text encoder $\tau$ are kept frozen. We only train a prompt prefix, termed Anonymization Prompt (AP), which is defined by
\begin{equation}
[AP] = [V_1][V_2]...[V_m].
\end{equation}
$[V_i]$ is a learnable vector for $i = 1, 2, \cdots m$, where $m$ is set as a constant. This prefix is prepended to any prompts that enter the diffusion model, which changes the text encoding function into 
\begin{equation}
\tau_{AP}(c) = \tau([AP]\oplus c),
\end{equation}
where $\oplus$ denotes a concatenation of the prompt and the tokenized text. As we will demonstrate, multiple advantages emerge in performing prompt learning instead of training entire models, including a plug-and-play property that enables transferring the prompt between different text-to-image models.

\subsection{Learning to Anonymize}
\label{sec:learning_anon}
Text-to-image diffusion models generate highly accurate facial images when prompted by names of public figures. A primary purpose of Anonymization Prompt Learning (APL) is to protect facial privacy by forcing the diffusion models to produce anonymized images, even when prompted to produce images of specific individuals. 

In existing moderation methods for diffusion models~\cite{erase, ablating, uce}, knowledge concerning target concepts are completely removed. However, for the case of face anonymization in text-to-image models, we endeavor to strike a balance between producing completely irrelevant images and retaining original facial attributes, following recent works in face anonymization~\cite{attranon}. In other words, when users request the images of a specific individual, we expect the model to return natural facial images with inaccurate identities, but share similar attributes to the requested identity. This can be especially valuable, as completely altering a person's appearance could introduce bias, misrepresentation, and degraded user experience. 

We reach the balance of anonymization by guiding the diffusion model with carefully curated text prompts, which force models to produce faces with similar facial attributes but distinct identities. We collect a specialized dataset $S_{ID} = \{(x^i, c_1^i, c_2^i)\}_{i=1}^N$ for APL. Each sample in dataset $S_{ID} $ is uniquely linked to an identity of a public figure. The sample triplet consists of an image of this person $x^i$, a prompt $c_1^i$ including the name of the person, and another prompt $c_2^i$ that describes the demographic information and prominent facial attributes of this person. This prompt is formatted as:
\begin{center}
prompt = A [gender] [race] [occupation], [facial attributes]. 
\end{center}
For example, for $c_1$ = "Albert Einstein", we have $c_2$ = "A male German-born physicist, bushy mustache, unkempt hair, deep-set eyes. " We note that prompting the model with $c_2$ alone almost never leads to the identity specified by $c_1$. Therefore, guiding the model to generate images of $c_2$ removes the identity, but lead to someone with similar facial attributes that fulfills the description of $c_2$. Concretely, we aim at aligning the conditional probabilities $P(x | c_1)$ and $P(x | c_2, \rm{not}\ c_1)$. Following existing work in compositional generalization~\cite{compositional_energy, compositional_diffusion}, the latter can be approximated with 
\begin{equation}
 P(x | c_2, \mathrm{not}\ c_1) \propto P(x) \frac{P(x | c_2)}{P(x | c_1)}.
\end{equation}

We compose the score function in such a way that approximates above probability:

\begin{equation}
\hat{\epsilon}_\theta (x, \tau(c_1), t) = \epsilon_\theta (x, \tau(c_2), t) + \alpha (\epsilon_\theta (x, \tau(c_2), t) - \epsilon_\theta (x, \tau(c_1), t)),
\end{equation}

where $\alpha$ is a weight parameter that controls the strength of the anonymization.

Finally, to enforce the maximization of the probability, we align the score functions modified by the Anonymization Prompt with this target score function:

\begin{equation}
L_{ID} = E_{(x, c_1, c_2)\sim S_{ID}, t}[||\epsilon_\theta(x, \tau_{AP}(c_1), t) -\hat{\epsilon}_\theta (x, \tau(c_1), t)||_2^2],
\end{equation}

Note that the gradients only flow through the first term with the Anonymization Prompt. Intuitively, $L_{ID}$ guides identifiable images produced by $c_1$ to move along the direction between $c_1$ and $c_2$ to align with the semantics of $c_2$, a non-identity-specific prompt. We empirically find this direction of anonymization to generalize very well, as models are able to generate anonymized identities for identities unseen during training.

\subsection{Quality Preservation}
\label{sec:sem_pres}

In practice, we expect text-to-image service providers to employ the Anonymization Prompt by prepending it to any prompts that enter the model. This way, no extra module is needed for examining the content of prompts. For identity-specific prompts, the Anonymization Prompt forces the model to automatically produce anonymized identities. To ensure practicality, it is critical that it does not degrade the base model's performance for non-identity-specific prompts, including those describing unspecified identities ("A person") or non-facial subjects ("An apple"). The aim is for APL to perform anonymization only when necessary, without compromising the model's ability to generate high-quality images across various prompts.

We reveal that given the formulation of $L_{ID}$ above, it is very simple to regularize Anonymization Prompt Learning for above purposes. Specifically, we adopt the LAION-2B~\cite{laion} dataset to curate a regularization dataset $S_{Reg} = \{(x^i, c_1^i, c_2^i)\}_{i=1}^N $ , which shares the data structure as $S_{ID}$. For each text-image pair $(c, x)$ in LAION, we curate a sample for $S_{Reg}$ by setting $x^i = x$ and $c_1 = c_2 = c$. We train on these samples with a similar loss function introduced above:

\begin{equation}
L_{Reg} = E_{(x, c_1, c_2)\sim S_{Reg}, t}[||\epsilon_\theta(x, \tau_{AP}(c_1), t) -\hat{\epsilon}_\theta (x, \tau(c_1), t)||_2^2].
\end{equation}

Effectively, we train the Anonymization Prompt to be aware of the prompts it is concatenated to. When prompts involve facial identities, the Anonymization Prompt enforces face anonymization. Otherwise, it is trained to be ineffective, as we guide the model to align the noise predictions for $\tau_{AP}(c)$ and $\tau(c)$. 

During training, we mix the samples of $S_{ID}$ and $S_{Reg}$ , and train the model with $L_{APL} = L_{ID} + L_{Reg}$. The pipeline of this process is presented in ~\cref{pipeline}.

\section{Experiments}
\label{sec:experiments}

In this section, we empirically demonstrate the effectiveness of Anonymization Prompt Learning (APL) in protecting facial privacy in the context of text-to-image models.

\subsection{Setup}
\label{sec:setup}
We first document details of training and evaluations for Anonymization Prompt Learning.

\textbf{Dataset.} We curate dataset $S_{ID} = \{(x^i, c_1^i, c_2^i)\}_{i=1}^N$ for training our Anonymization Prompt. We first collect names of identities that encompass different genders, races, and occupations. With these names as prompts, training images $x^i$s are generated with Stable Diffusion v1.5~\cite{ldm}. We use ChatGPT~\cite{chatgpt} to generate $c_2$ prompts by filling in names and requesting descriptions of the individual's demographic characteristics and prominent facial attributes. While these responses potentially include inaccurate descriptions, we find this experimental collection procedure to be quite effective in our experiments. We additionally collect another dataset of identities $S_{test}$ for evaluating anonymization on unseen identities. Both $S_{ID}$ and $S_{test}$ contain 100 non-overlapping identities of public figures. Samples of these datasets can be found in the Appendix. 

\textbf{Training}. The Anonymization Prompt is trained on the base model of Stable Diffusion v1.5~\cite{ldm} (SD-v1.5), an advanced text-to-image diffusion model. Hyperparameters are set as $m = 10$ and $\alpha = 1$. During training, we set the learning rate at $1e-3$, and train for $20,000$ steps with batch sizes of 1. We keep all model parameters frozen and only train the learnable prompt. 

\textbf{Evaluating Anonymization}. 
We measure the identity accuracy (ID-ACC) of generated images by calculating the cosine similarity between its Arcface~\cite{arcface} face recognition features and that of the ground truth identity. Lower identity accuracies imply better anonymization. Besides evaluating on SD-v1.5, we demonstrate the transferability of our method with three other text-to-image models, namely SD-v1.4~\cite{ldm}, RealisticVision v5.1~\cite{realisticvision}, and Stable Diffusion XL (SDXL)~\cite{sdxl}. Note that SDXL~\cite{sdxl} utilizes a text encoder with different sizes of text embeddings, which bars a direct transfer of our soft prompt. We resolve this issue with a zero-shot prompt transfer method proposed in~\cite{prompt-transfer}, mapping our prompt to a different embedding space, so that our method remains effective with different model architectures.

\textbf{Evaluating Quality}. We assess the generation quality of both identity-specific and non-identity-specific images. For the former, we calculate FID~\cite{cleanfid} and facial attribute similarities between faces generated by our model and the original model. For the latter, we generate images with captions from the MS COCO 2014 evaluation set~\cite{coco}, following common practices in generative modeling research~\cite{ldm, imagen, dalle3}. We report CLIP~\cite{clip} and FID~\cite{cleanfid} scores to respectively evaluate semantic and quality preservation. Given that our method modifies an existing generative model, we calculate FID with reference images generated by the original model in all cases.

\subsection{Anonymization across Models}
\label{sec:anonacrossmodels}
\begin{table}[ht]
\centering
\vspace{-0.5em}
\caption{The effect of the Anonymization Prompt on identity accuracies of three text-to-image models. All models generate highly accurate identities when prompted with names of public figures. Employing the prompt significantly reduces accuracies. We calculate \textbf{mean} / \textbf{maximum} identity accuracies for images generated from the same identity, and aggregate these metrics across all identities.}
\label{tab:id_acc}
\begin{tabular}{@{}l @{\hspace{20pt}}c @{\hspace{20pt}}c@{}}
\toprule
\multirow{2}{*}{Models} &  \multicolumn{2}{c}{Identity Accuracy (mean / max)}\\ 
\cmidrule{2-3}
            ~ & Training set ($S_{ID}$) & Testing set ($S_{test}$) \\ \midrule
SD-v1.5~\cite{ldm}              & 0.31566 / 0.37958       & 0.30254 / 0.36691       \\
SD-v1.5~\cite{ldm} + APL       & 0.03924 / 0.10861       & 0.03970 / 0.10773     
\\\midrule
SD-v1.4~\cite{ldm}               & 0.29911 / 0.36713       & 0.28837 / 0.35720        \\
SD-v1.4~\cite{ldm}  + APL       & 0.03893 / 0.10632       & 0.04647 / 0.11368   \\\midrule  
RV-v5.1~\cite{realisticvision}                    & 0.32205 / 0.37122       & 0.32071 / 0.36492       \\
RV-v5.1~\cite{realisticvision} + APL               & 0.09215 / 0.14391       & 0.09033 / 0.14496       \\\midrule
SDXL~\cite{sdxl}         &    0.27917 / 0.33371    & 0.27218 / 0.32830
       \\
SDXL~\cite{sdxl} + APL    &    0.07267 / 0.14245    &  0.07852 / 0.14368

  \\\bottomrule
\end{tabular}
\end{table}

\begin{figure}[t]
  \centering
  \includegraphics[width=1\linewidth]{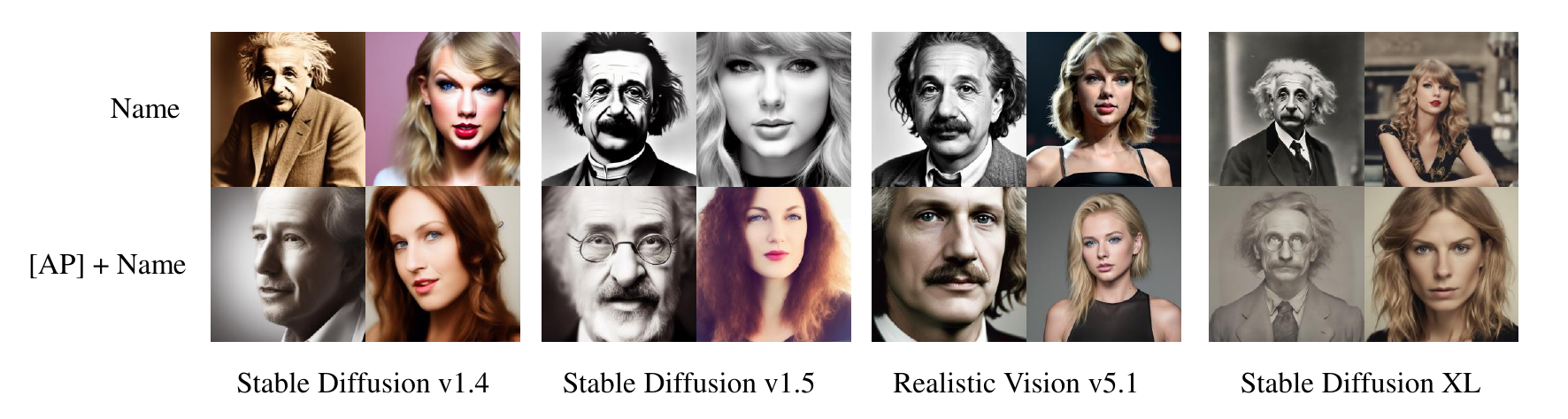}
  \caption{Images generated by three different models when employing the same Anonymization Prompts (AP). Models are prompted to generate Albert Einstein and Taylor Swift, which respectively belong to the training set and the testing set. Note that the AP is only trained on SD-v1.5~\cite{ldm}. The visual effect demonstrate successful anonymization for any identities across different text-to-image models, consistent with our quantitative results.  }
  \label{fig:fig3}
  
\end{figure}

Following the training of the Anonymization Prompt, we prepend the learned prefix to different text-to-image diffusion models to evaluate its anonymization performances. We prompt the models to generate images of public figures and calculate identity accuracies (ID-ACC). In this experiment, we use the prompt template "A close-up portrait of [Name]" to ensure that the models generate clear depictions of faces. For each identity, we generate a group of images and calculate both mean and maximum ID-ACCs within the group. Subsequently, we aggregate these scores across all identities to present overall mean and maximum identity accuracies. Results are presented in~\cref{tab:id_acc}. 

Experiments show that original text-to-image models perform well on generating facial identities, reaching an average ID-ACC of around 0.3. However, with the Anonymization Prompts, ID-ACCs drop to as low as 0.039 with SD-v1.5, marking almost a tenfold decrease from the original version. Comparable results of less than 0.1 are obtained on three other models, which is sufficient to render the identities unrecognizable. Remarkably, the anonymization performances on the unseen testing set are on par with those on the training set, indicating an effectiveness to all identities, beyond those available during training. 

Visually, faces generated by all models are significantly less recognizable, as demonstrated in ~\cref{fig:fig3}, consistent with our quantitative results. We thereby confirm the validity of the Anonymization Prompt on four different models, including the state-of-the-art SDXL~\cite{sdxl}. This showcases a remarkable transferability of our method, given the prompt is only trained on SD-v1.5.

\subsection{Quality Preservation}
\label{sec:semantic}

We strive to ensure our anonymization is natural and specific to identifiable facial images. When requested to generate specific individuals, models equipped with the Anonymization Prompts should return realistic facial images with false identities, but retain facial attributes similar to those of the requested identity, as outlined in \cref{sec:learning_anon}. In addition, this modification should not interfere with generation qualities of unrelated images, \emph{i.e.}, concatenating the Anonymization Prompt to prompts of non-facial images and non-identity-specific facial images should not significantly change the outputs. 

In this section, we comprehensively evaluate the ability of our method to meet these requirements. We first study the quality of 4000 facial images generated with identity-specific prompts from our dataset $S_{test}$. We calculate the FID~\cite{cleanfid} score of our generated faces against those generated by the original model. For attribute evaluation, we calculate the Attribute Accuracy (Attr-ACC) of generated facial images, which is the average cosine similarity between the facial attribute features~\cite{farl} of the generated faces and the ground truths. For non-identity-specific generations, we generate $30k$ images with prompts from COCO-30k~\cite{coco}. We evaluate FID scores between images generated by modified models and the original Stable Diffusion~\cite{ldm}, and evaluate CLIP scores by averaging the distances between all generated images and their corresponding text prompts. 

A direct baseline of our task is to apply concept editing methods~\cite{erase, ablating, forgetmenot, uce} and remove facial identities as target concepts. We compare our results with state-of-the-art concept editing, namely ESD~\cite{erase}, Concept Ablation (CA)~\cite{ablating}, UCE~\cite{uce} and Forget-me-not (FMN)~\cite{forgetmenot}. We run their open training code on SD-v1.5~\cite{ldm} but substitute their editing concepts with identities in our training set.

\begin{table}[ht]
\centering
\vspace{-0.4em}
\caption{Quantitative results of quality preservation. We report performance on ID-specific generation and non-ID-specific generation, which are respectively evaluated with ID-specific prompts and prompts from COCO-30k~\cite{coco}.  Best results are in \textbf{bold} and second bests are \underline{underlined}. }

\label{tab:fidclip}
\begin{tabular}{l|ccc|cc}
    \toprule
        ~ & \multicolumn{3}{c|}{ID-specific} & \multicolumn{2}{c}{Non-ID-specific}\\ \hline
        ~ & FID (↓) & ID-ACC (↓) & Attr-ACC (↑) & FID (↓) & CLIP (↑)  \\ \hline
        SD-v1.5~\cite{ldm} & - & 0.30254 & 0.69361 & - & 0.3078  \\ \hline
        FMN~\cite{forgetmenot} & 263.57 & - & - & 54.89 & 0.1642  \\ 
        UCE~\cite{uce} & 125.67 & - & - & 88.69 & 0.0298  \\ 
        ESD~\cite{erase} & 189.32 & - & - & 9.59 & 0.2530  \\ 
        CA~\cite{ablating} & \underline{57.58} & \textbf{0.01215} & \underline{0.53898} & \underline{4.14} & \underline{0.2903}  \\ 
        Ours (APL) & \textbf{20.01} & \underline{0.03970} & \textbf{0.60222} & \textbf{2.98} & \textbf{0.3058}  \\ \bottomrule
    \end{tabular}
\end{table}

Results in~\cref{tab:fidclip} demonstrate that our method best satisfies all requirements for a successful anonymization. It not only generates high quality facial images with low identity accuracies, but also preserves facial attribute information of the requested identities. On generating COCO~\cite{coco} images, APL achieves the highest image quality and text fidelity, respectively quantified by FID and CLIP scores, surpassing all compared methods.

In contrast, the compared methods fail to meet the demand of our anonymization task. After training with our dataset, some methods exhibit difficulties in generating detectable faces, therefore we do not report their ID-ACC and Attr-ACC. For example, FMN~\cite{forgetmenot} and UCE~\cite{uce} produce high FID scores when prompted with specific identities, such that a large number of images do not contain any faces. Simultaneously, they generate COCO images with high FID scores and low CLIP scores, as the resulting images are neither realistic nor related to the input prompts. This is outcome is consistent with existing research~\cite{uce}, which suggests that editing too many concepts with existing methods may significantly impact the generation quality. 

\begin{figure}[t]
  \centering
  \includegraphics[width=1\linewidth]{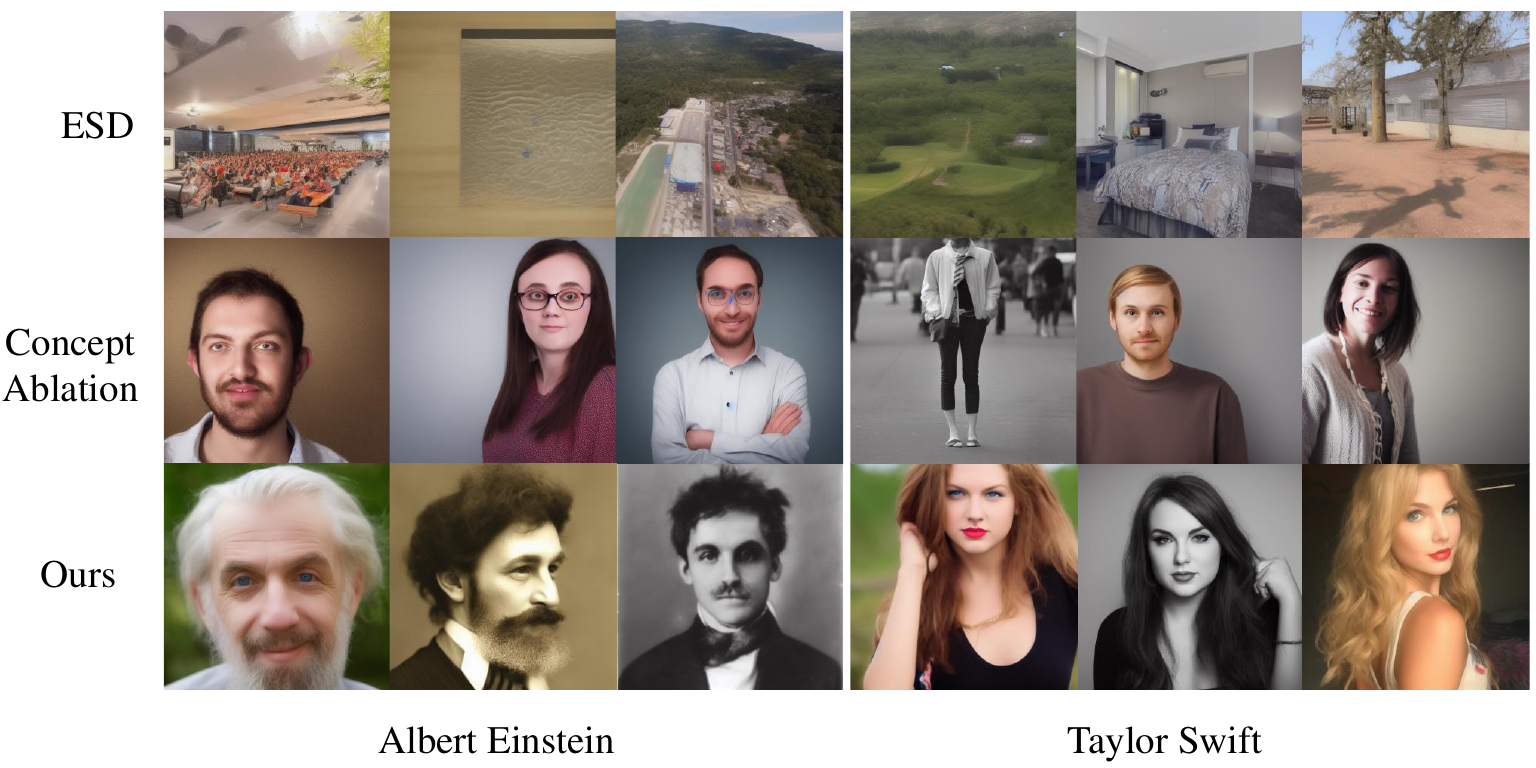}
  \caption{Excessive anonymization observed in compared methods. Empty string targets of ESD~\cite{erase} lead to random scenery generations and category label targets of Concept Ablation~\cite{ablating} generate every identity as a random person. Our prompts curated with abundant attribute information strikes a balance between erasure of semantics and attaining original attributes. }
  \label{fig:vis_excessive}
\end{figure}

Meanwhile, we show that some methods fail to meet the requirements despite relatively maintaining generation quality. For example, models trained by ESD~\cite{erase} generates random scenery images when prompted with specific identities, as shown in~\cref{fig:vis_excessive}. This occurs because ESD~\cite{erase} aligns its target concepts to empty strings, effectively removing the concepts entirely. Similarly, Concept Ablation~\cite{ablating} selects anchor concepts, such as "a person", as optimization targets. Although it generates facial images with low identity accuracies, this is at the cost of minimal facial attribute similarity with the requested identity, as indicated by a markedly lower Attr-ACC. As discussed in~\cref{sec:learning_anon}, we consider such anonymization excessive, as they produce completely irrelevant faces or no faces at all. We visualize excessive anonymization in~\cref{fig:vis_excessive}.

APL sets appropriate optimization targets that specifies basic demographic and facial attribute information for all training identities, and therefore preserves image quality and necessary attributes while removing identity information. Its regularization loss $L_{Reg}$ effectively preserves generation quality of unrelated images, as we show in the ablation study. Finally, its prompt learning setting not only ensures minimal quality degradation compared to the original model, but also makes it the only method that can be trained once and then transferred to other models.

\subsection{Anonymizing New Identities}
\label{sec:personalization}
Personalization methods, such as Dreambooth~\cite{dreambooth}, Textual Inversion~\cite{textualinversion} and HyperDreambooth~\cite{hyperdreambooth}, enable diffusion models to acquire new concepts, including highly accurate facial identities~\cite{hyperdreambooth}. This capability significantly expands deepfake threats, affecting individuals beyond those in the model's training dataset. Meanwhile, these methods also undermine the effectiveness of traditional concept editing methods~\cite{erase, ablating}, as studies show that concepts removed by them can be re-introduced to the model with the same personalization procedure~\cite{imma, ringabell}.

Our experiments demonstrate that our method remains effective amid these challenges. Firstly, it effectively anonymizes identities learned with Dreambooth, even though the identities was never seen during training. Secondly, because APL does not modify model parameters, re-learning anonymized identity is not effective. After any attempts of re-learning, users can simply add the prompt back on and still generate anonymized images. These capabilities are not preserved by previously compared methods, which unanimously modify model parameters.

\begin{figure}[h]
  \centering
  \vspace{-1em}
  \includegraphics[width=1\linewidth]{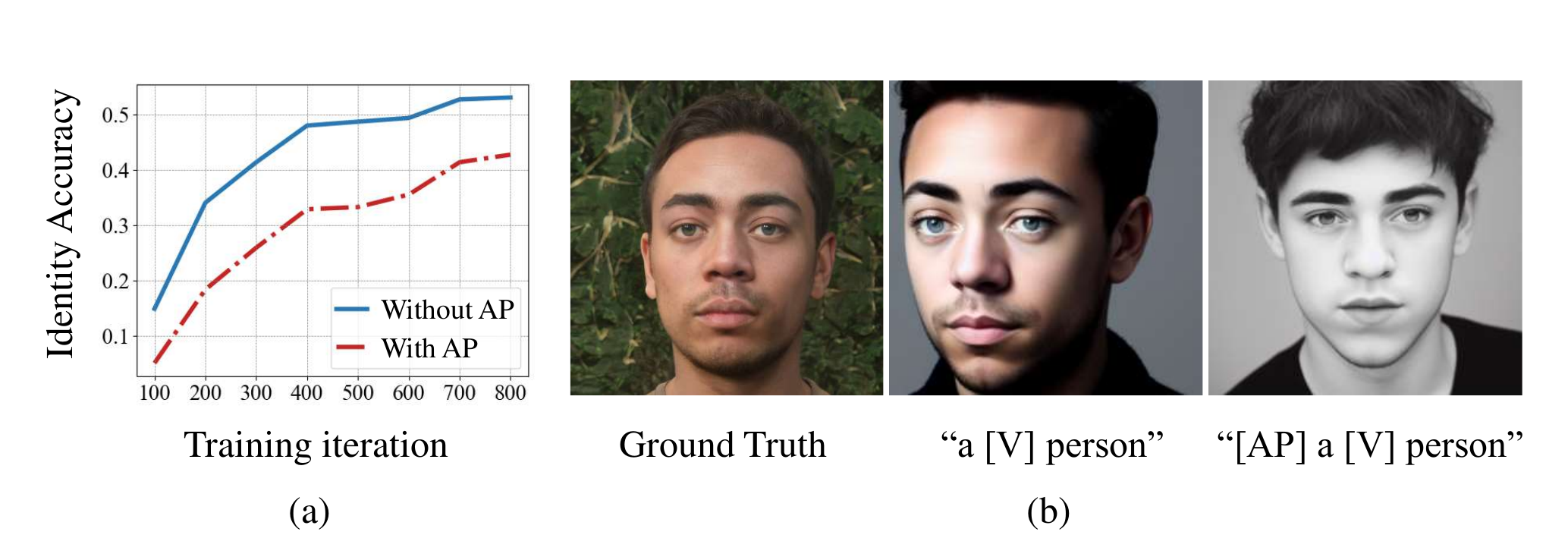}
  \caption{We evaluate the anonymization performance of our Anonymization Prompt (AP) on identites learned after APL training. (a) Identity accuracies of images generated with and without the Anonymization Prompt at different Dreambooth~\cite{dreambooth} iterations. The prompt consistently lowers identity accuracies on newly-learned identities. (b) An example of anonymizing a new identity~\cite{hyperdreambooth}, which is linked to a token [V] by Dreambooth~\cite{dreambooth}. The prompt significantly reduces the identity accuracy of the generated image.  }
  \label{dreambooth}
\end{figure}

In specific, we introduce new identities to Stable Diffusion v1.5~\cite{ldm} with Dreambooth~\cite{dreambooth} and evaluate the anonymization performance of the Anonymization Prompt on these identities. We sample 50 identities from the VGGFace-2~\cite{vggface2} dataset, collecting high-resolution photos for each identity for personalization. We obtain a personalized Stable Diffusion for each of those identities with Dreambooth~\cite{dreambooth}, so that resulting models are able to accurately generate their corresponding identities with a pre-determined token. Models are fine-tuned for 800 iterations, which is sufficient for the base model to learn accurate identities. During training, we generate images of the corresponding identities every 100 iteration, both with and without applying the Anonymization Prompt, to measure their identity accuracies. The results across all identities are then aggregated to form the curves in~\cref{dreambooth}(a).

Owing to the effectiveness of Dreambooth~\cite{dreambooth}, both curves exhibit a clear increasing trend. However, in cases where identities are generated with the Anonymization Prompt, the identity accuracies consistently remain lower compared to those generated without the Anonymization Prompt across all iterations. With fewer iterations, the prompt effectively reduces the identity accuracies to nearly half of the base model. At 800 iteratieons, generations with the prompt achieves an average identity accuracy of 0.43, markedly lower than the baseline of 0.53, demonstrating effective anonymizations on these new identities when Dreambooth~\cite{dreambooth} is trained until convergence. We visualize the effect of the Anonymization Prompt on one learned identity in~\cref{dreambooth}(b). At the same training iteration of Dreambooth~\cite{dreambooth}, the identity generated without the prompt closely resembles the ground truth identity, whereas the one generated with the prompt is much less recognizable, showcasing successful anonymization.

These figures demonstrate both quantitatively and qualitatively that our Anonymization Prompt effectively anonymizes identities learned by the base model after the initial prompt learning. Consequently, modifications to the model parameters to either introduce new identities or restore previously erased ones do not affect our anonymization performance. This finding significantly extends the applicability of APL to encompass all possible facial identities, thereby offering protection against deepfake threats and privacy violations not only for public figures but also for the general population.

\subsection{Ablation Study}
\label{sec:ablationstudy}

We study the effects of several key design choices and hyperparameter selections for Anonymization Prompt Learning, including training iterations, regularization, prompt length, weight parameter $\alpha$ and the size of the training dataset. 

\begin{figure}[h]
  \centering
  \vspace{-1em}
  \includegraphics[width=1\linewidth]{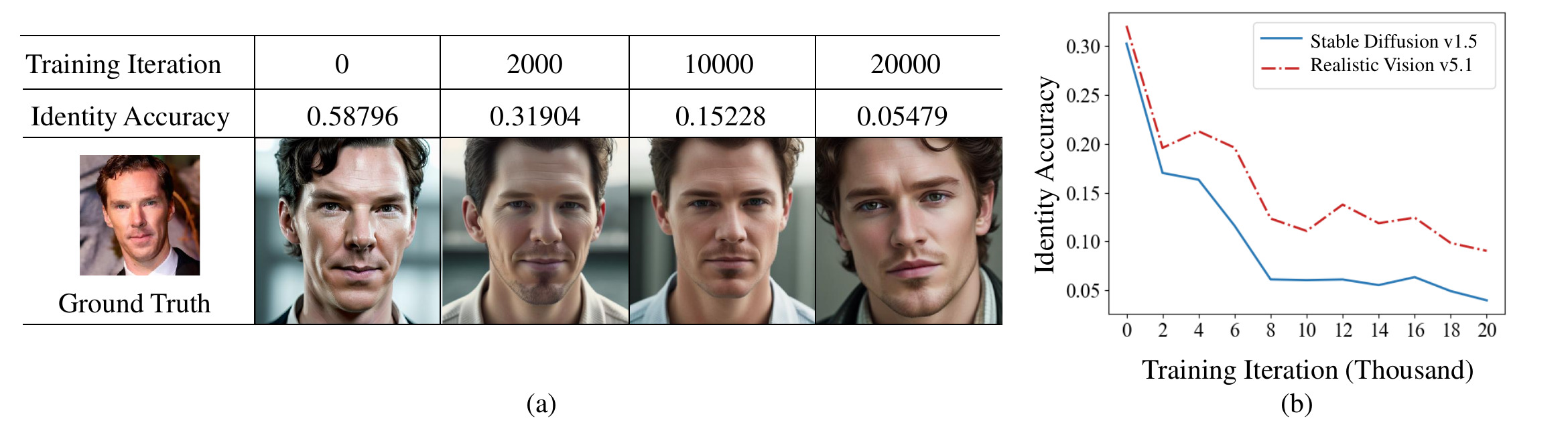}
  \caption{Effects of training iterations. (a) An identity generated by RealisticVision~\cite{realisticvision} that employs the Anonymization Prompts of different training iterations. Different anonymization levels of the same identity is observed. (b) Identity accuracies computed across the test set. Users of our method may choose to use the Anonymization Prompts at different training iterations according to their own policies of content moderation. }
  \label{iterations}
\end{figure}

\textbf{Training iterations}.
We monitor the identity accuracies during the entire training process, and observe that the accuracies for all models progressively decrease over time in~\cref{iterations}(a). This effect is visualized with one identity in~\cref{iterations}(b). Both visual effects and quantitative measures indicate a steadily increasing level of anonymization. In light of these findings, text-to-image service providers may choose to employ the Anonymization Prompts at various stages of training to enforce different levels of anonymization in alignment with their content moderation policies. 

\begin{figure}[h]
  \centering
  \includegraphics[width=1\linewidth]{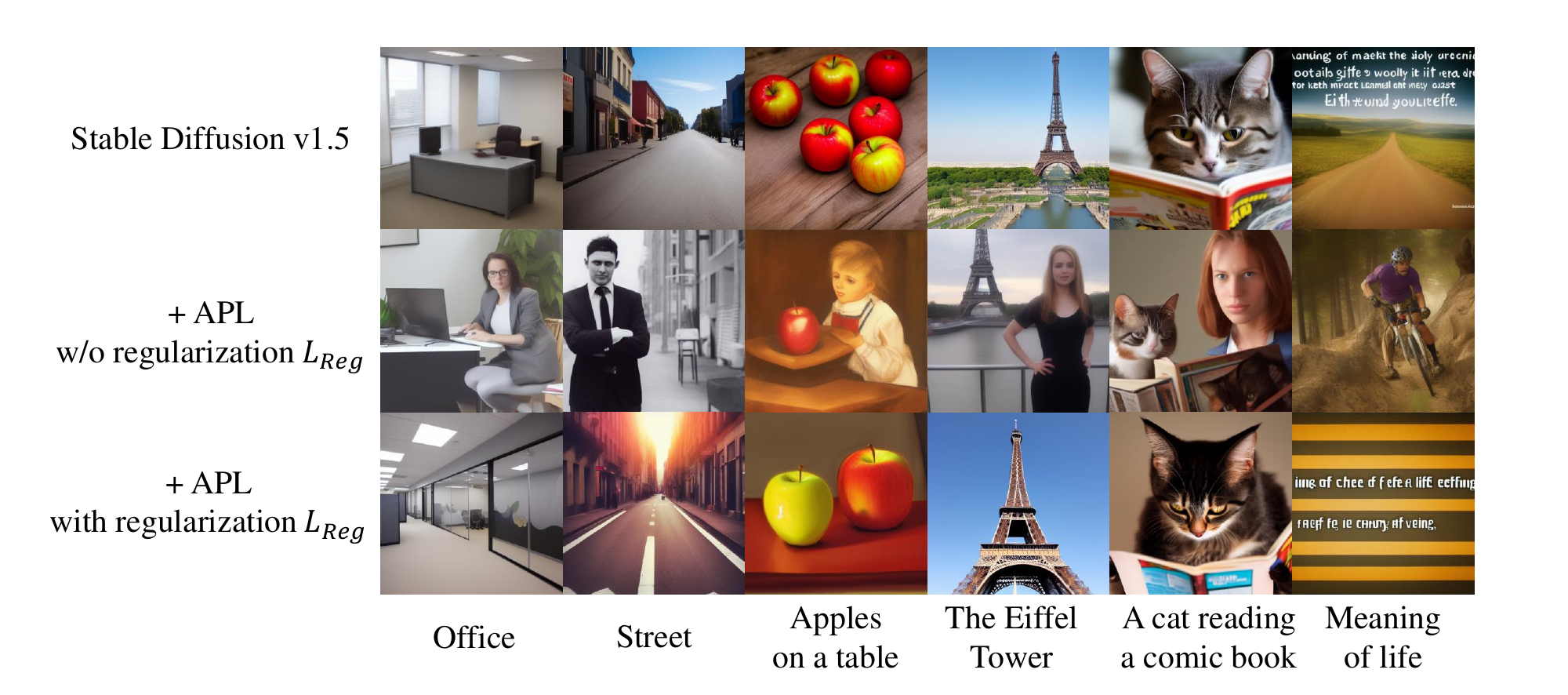}
  \vspace{-1em}
  \caption{Effect of our regularization loss $L_{Reg}$. When training APL solely with facial images, the model links the prompt to the concept of faces, leading to unexpected depictions of people in every generation. Regularization removes the issue by inactivating the Anonymization Prompt for non-ID prompts. }
  \label{reg}
  
\end{figure}

\textbf{Regularization}.
We ablate the regularization loss $L_{Reg}$, which trains the Anonymization Prompt to be ineffective when concatenated to non-identity-specific prompts. We demonstrate in~\cref{reg} that without regularization, models generate unexpected depictions of people, which is a significant violation of semantic alignment. Since APL is a prompt learning method, training without the regularization of non-face images causes the model to link the prompt to the concept of faces, resulting in above situations. Therefore, training the prompt to ensure its ineffectiveness on non-identity-specific prompts is essential for our method to be useful in practice.

\renewcommand{\arraystretch}{1.2}
\begin{table}[!ht]
    \centering
    \caption{Identity Accuracy of the Anonymization Prompt with different prompt length $m$. We report \textbf{mean} / \textbf{maximum} ID-ACC in a group of image for the same individual, and then aggregate across all identities. }
    \label{tab:prompt_length}
    \begin{tabular}{@{}c @{\hspace{20pt}}c @{\hspace{20pt}}c@{}}
    \toprule
        \multirow{2}{*}{Prompt length} &  \multicolumn{2}{c}{Identity Accuracy (mean / max)}\\ 
        \cmidrule(lr){2-3}
        ~ & Training set & Testing set\\
        
        \midrule
        1 & 0.12308 / 0.20149 & 0.11434 / 0.19768 \\ 
        5 & 0.05834 / 0.12377 & 0.05758 / 0.13179 \\ 
        10 & 0.03345 / 0.10548 & \textbf{0.03257 / 0.09702} \\ 
        20 & \textbf{0.03270 / 0.09735} & 0.04084 / 0.10780 \\ \bottomrule
    \end{tabular}
\end{table}

\textbf{Prompt length}. The length of the Anonymization Prompt is the number of learnable token that constitutes our prompt prefix. We study the optimal length of the prompt by evaluating Identity Accuracies with prompts of different lengths. As shown in ~\cref{tab:prompt_length}, we find that $m=10$ strikes a balance between effective anonymization and computational efficiency. Shorter prompts are less effective at anonymization. Longer prompts only offer marginal improvements, while requiring more computational and storage resources.

\renewcommand{\arraystretch}{1.2}
\begin{table}[h]
    \centering
    \caption{Identity Accuracy of the Anonymization Prompt trained with different weight parameter $\alpha$. We report \textbf{mean} / \textbf{maximum} ID-ACC in a group of image for the same individual, and then aggregate across all identities. }
    \label{tab:eta}
    \begin{tabular}{@{}r @{\hspace{20pt}}c @{\hspace{20pt}}c@{}}
    \toprule
        \multirow{2}{*}{$\alpha$} &  \multicolumn{2}{c}{Identity Accuracy (mean / max)}\\ 
        \cmidrule(lr){2-3}
        ~ & Training set & Testing set\\
        \midrule
        0 & 0.05237 / 0.12196 & 0.05688 / 0.13048 \\ 
        0.5 & 0.04826 / 0.11933 & 0.04090 / 0.10900 \\ 
        1 & \textbf{0.03345 / 0.10548} & \textbf{0.03257 / 0.09702}  \\ 
        2 & 0.06819 / 0.14412 & 0.06707 / 0.14137 \\ 
        5 & 0.05622 / 0.13619 & 0.04083 / 0.10900 \\ \bottomrule
 \end{tabular}
\end{table}

\textbf{Weight parameter $\alpha$}. The weight parameter $\alpha$ controls the strength of modification. We empirically find $\alpha=1$ to provide the best anonymization. Intuitively, $\alpha=0$ implies precisely aligning identifiable facial images with a prompt without identity specifications. A larger $\alpha$ implies going further along the direction of anonymization, thus leading to lower ID-ACC. However, with $\alpha > 1$, the training process may become more sensitive to noise in the estimation of the anonymization direction, thereby adversely affecting the results. 

\begin{figure}[h]
  \centering
  \includegraphics[width=0.8\linewidth]{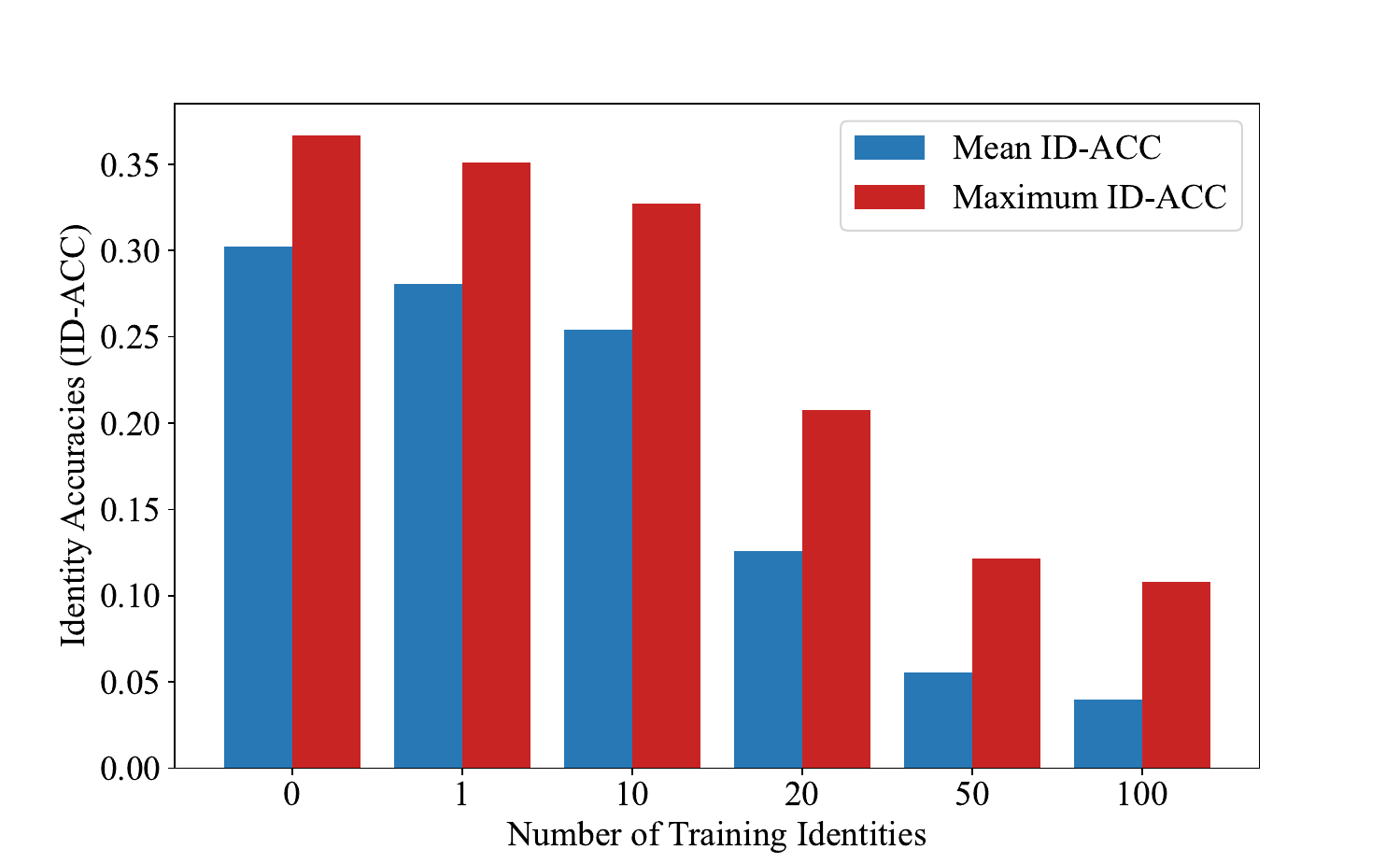}
  \caption{We evaluate the identity accuracies (ID-ACC) of Stable Diffusion v1.5, enhanced with the Anonymization Prompts (AP) trained on differing sizes of datasets. Zero denotes evaluations without the Anonymization Prompt. Evaluation is performed on the test set $S_{test}$ distinct from training identities. We document both the \textbf{mean} and \textbf{maximum} ID-ACC for a group of images per identity, then aggregate these metrics across all identities. We observe a consistent decline in both mean and maximum ID-ACC as the number of identities in the training set increases.  }
  \label{num_id}
  \vspace{-1em}
\end{figure}

\textbf{Dataset Size. } The Anonymization Prompt enables the anonymization of any identities, including those unseen during training and those learned through fine-tuning. We show that this generalization ability has its roots in the number of training identities. Results are presented in ~\cref{num_id}. With more training identities, we observe a consistent decrease in both average and maximum Identity Accuracies. This suggests that incorporating more identities enables the model to identify more precise directions from $c_1$ to $c_2$ in the latent space, where $c_1$ represents the prompts specifiying identities and $c_2$ are prompts encapsulating the demographic information and facial attributes of these identities. An accurate estimation of this direction guides the model to anonymize more identities successfully. We expect the identity accuracies to further drop on unseen diffusion models if larger datasets with more diverse attributes are available.

\vspace{-0.5em}
\section{Conclusions}
\label{sec:conclusions}
\vspace{-0.5em}
We propose Anonymization Prompt Learning (APL), which produces a plug-and-play Anonymization Prompt that can be inserted to any text-to-image diffusion models and enforce the generation of anonymized faces. We show the ability of the Anonymization Prompt to anonymize various identities on different models while preserving the basic generation performances of the original diffusion model. Put together, our method has the potential to be incorporated by text-to-image service platforms to prevent the generation of identity-specific images and address deepfake generations. We hope the development of the method contributes to the long-term battle against malicious use of AI-generated content. 

\newpage
\begin{appendices}

\section{Samples from collected dataset}

We present some samples from our collected dataset $S_{ID}$ in~\cref{tab:samples}. We reiterate that the attribute information is automatically generated by ChatGPT~\cite{chatgpt}, and they may not accurately reflect the actual characteristics of these individuals.

\renewcommand{\arraystretch}{1.5}
\begin{table}[h]
    \caption{Six samples from our collected dataset $S_{ID}$. We collect a diverse set of individuals and generate their attribute description with ChatGPT~\cite{chatgpt}.}
    \label{tab:samples}
    \centering
    \begin{tabular}{ll}
    \hline
        Names ($c_1$)  & Emma Stone  \\ \hline
        Attributes ($c_2$)  & A female American actor, green eyes, vibrant smile  \\ \hline
        Names ($c_1$)  & David Beckham  \\ \hline
        Attributes ($c_2$)  & A male British soccer player, blond hair, tattoos.  \\ \hline
        Names ($c_1$)  & Jackie Chan  \\ \hline
        Attributes ($c_2$)  & A male Chinese actor and martial artist, short hair, athletic build and expressive face  \\ \hline
        Names ($c_1$)  & Britney Spears  \\ \hline
        Attributes ($c_2$)  & A female American singer and actor, blond hair, brown eyes, pop icon look  \\ \hline
        Names ($c_1$)  & Steve Jobs  \\ \hline
        Attributes ($c_2$)  & A male American entrepreneur, black turtleneck, round glasses, short, grey hair  \\ \hline
        Names ($c_1$)  & Albert Einstein  \\ \hline
        Attributes ($c_2$)  & A male German-born physicist, bushy mustache, unkempt hair  \\ \hline
        
    \end{tabular}
\end{table}

The collection of this dataset can be straightforwardly reproduced by querying ChatGPT~\cite{chatgpt} with a list of celebrity names, with prompts formatted as follows: 

\begin{quote}
\texttt{Here is a list of famous individuals. Using your knowledge, please generate a concise, one-sentence description for each person that includes their gender, ethnicity, profession, and notable facial features. For instance, for the name Albert Einstein, you would provide: "A male, German-born physicist with a bushy mustache and unkempt hair." [List of names]}
\end{quote}

The specific prompt can be varied for generating similar data.




\end{appendices}

\newpage
\bibliography{sn-bibliography}

\end{document}